\documentclass[journal]{IEEEtran}

\usepackage{amsmath}
\usepackage{amssymb}
\usepackage{graphicx}
\usepackage{booktabs}
\usepackage{geometry}
\usepackage{url}
\usepackage[colorlinks=true, linkcolor=black, citecolor=black, urlcolor=black]{hyperref}

\ifCLASSINFOpdf
\else
\fi

\begin{document}

\title{Unsupervised Pattern Analysis in Japanese Veterinary Toxicology: A Regulatory-Compliant Framework for Cross-Species Risk Assessment}

\author{
\IEEEauthorblockN{
Yukiko Kawakami\textsuperscript{1},
Mohammad Shirazi\textsuperscript{2},
Ryo Shimizuwa\textsuperscript{1},
Saito Shinoda\textsuperscript{3},
Alireza Mortazavi\textsuperscript{2},
Matsumoto Kawahara\textsuperscript{1,*}
}\\\vspace{0.5em}
\IEEEauthorblockA{\textsuperscript{1}Tohoku University Graduate School of Medicine, Sendai, Japan}\\
\IEEEauthorblockA{\textsuperscript{2}Amirkabir University of Technology, Tehran, Iran}\\
\IEEEauthorblockA{\textsuperscript{3}Tohoku University Graduate School of Pharmaceutical Sciences, Sendai, Japan}
\thanks{*Corresponding Author: mkawahara@grp.tohoku.ac.jp}
}

\markboth{IEEE Transactions on Biomedical Engineering}%
{Shell \MakeLowercase{\textit{et al.}}: Bare Demo of IEEEtran.cls for IEEE Journals}

\maketitle

\begin{abstract}
Veterinary pharmacovigilance systems are essential for monitoring adverse drug events (ADEs), yet existing approaches often fail to capture region-specific toxicity patterns shaped by local biological and regulatory contexts. In Japan, these challenges are amplified by species-specific metabolic differences and reporting practices defined by the Ministry of Agriculture, Forestry, and Fisheries (MAFF). Most prior work relies on prediction-oriented models, limiting mechanistic interpretability. This study proposes a regulatory-integrated unsupervised framework for pattern discovery using the National Veterinary Assay Laboratory (NVAL) database. ADEs are encoded into organ system-aligned representations and adjusted for species-specific reporting biases, enabling cross-species comparison. Similarity-based clustering and dimensionality reduction are applied to identify latent toxicity structures. Analysis of 4,120 high-confidence ADE reports (9,080 drug-ADE combinations) identified three significant species clusters (p $<$ 0.01), including hepatic-dominant patterns in companion animals (0.42 ± 0.06), renal toxicity in ruminants (0.39 ± 0.07), and dermatological sensitivity in sheep (0.35 ± 0.07). Drug-level clustering achieved 83\% alignment with pharmacological classes, while cosine similarity outperformed alternative metrics (silhouette score: 0.48; cluster precision: 87\%). Regulatory validation showed strong agreement with established classifications. These findings demonstrate that regulation-aligned unsupervised analysis can uncover biologically meaningful, region-specific toxicity patterns, providing an interpretable and scalable framework for veterinary drug safety assessment.
\end{abstract}


\IEEEpeerreviewmaketitle

\section{Introduction}
Veterinary pharmacovigilance systems globally monitor adverse drug events (ADEs) to safeguard animal health, yet current approaches often fail to capture region-specific toxicological vulnerabilities \cite{mekasha2024narrative}. In Japan, this limitation is particularly acute: a substantial proportion of ADE reports remain unexplained by existing predictive models, while regulatory guidelines, such as the Ministry of Agriculture, Forestry and Fisheries (MAFF) guidance \cite{maff_2023}, explicitly prioritize species-specific metabolic pathways over universal predictive rules \cite{konno2022insights}. This gap arises because dominant toxicology frameworks, largely developed from Western data, disregard critical regional variables: Japanese agricultural practices (e.g., extensive sheep-rearing in Hokkaido \cite{nagai2022temporal}), species-specific metabolic variations (e.g., feline CYP3A2 deficiency limiting steroid metabolism \cite{shah2007characterization}), and national reporting biases (e.g., higher renal event documentation in cattle than dogs \cite{hutchinson2014comparative}).

Previous studies in veterinary pharmacovigilance have largely focused on prediction-oriented tasks, such as classifying adverse event severity or estimating mortality risk \cite{sholehrasa2025predictive, zhang2023analysis, yan2021severity}. While effective for specific outcomes, these approaches treat ADEs as isolated endpoints, overlooking the underlying biological and species-specific mechanisms that shape toxicity patterns. This limitation is particularly evident in Japan, where data from the National Veterinary Assay Laboratory (NVAL) Veterinary Drug Side Effects Database \cite{nval_2023} reflect distinct metabolic profiles, reporting practices, and regulatory priorities that are not captured by globally trained models \cite{konno2022insights}.

To address these limitations, this study adopts a pattern-discovery perspective, focusing on identifying latent toxicity structures rather than predicting outcomes. We propose a regulatory-integrated unsupervised framework that encodes ADEs using Japan's NVAL Veterinary Drug Side Effects Database (NVAL database)-aligned representations, while incorporating species-sensitivity considerations derived from MAFF guidelines \cite{maff2024_report}. By aligning computational analysis with regulatory standards and biological context, this approach enables systematic exploration of cross-species toxicity relationships in Japanese pharmacovigilance data.

By prioritizing biological interpretability and regulatory alignment, the proposed framework enables the identification of region-specific toxicity patterns that are not observable in conventional models. This shift from prediction to structured pattern discovery provides a foundation for more targeted and context-aware veterinary safety assessment in Japan.

\section{Literature Review}

Veterinary pharmacovigilance research has evolved through two dominant paradigms: prediction-oriented classification and regulatory-compliant surveillance. While these approaches have advanced global drug safety monitoring, they exhibit critical limitations in region-specific contexts such as Japan, where species biology, agricultural practices, and regulatory frameworks create distinct toxicity landscapes.

\subsection*{Prediction-Centric Approaches}

Early veterinary toxicology studies prioritized machine learning prediction for adverse event outcomes, typically trained on Western datasets such as the FDA Adverse Event Reporting System \cite{faers}. Liu et al. developed a deep learning model to predict canine mortality from NSAID use \cite{liu2025predicting}, yet its reliance on US reporting patterns limits its applicability in Japan, where mortality accounts for fewer ADE reports. Similarly, Shah et al. \cite{shah2007characterization} classified feline renal toxicity without accounting for Japan-specific metabolic factors, such as \textit{CYP3A2} deficiency.

Beyond dataset bias, these approaches are fundamentally outcome-centric, framing toxicity as a binary or coarse categorical prediction task (for example, death vs. non‑death or toxic vs. non‑toxic), which simplifies complex temporal and mechanistic responses into outcome labels that are easy to predict but can obscure nuanced risk patterns \cite{eduati2015prediction}. This obscures the continuous and organ-specific nature of adverse effects. Recent work has further reinforced this prediction-oriented paradigm by developing high-performing models to classify animal health outcomes (e.g., death vs. recovery), with a strong emphasis on predictive accuracy and recall for fatal events \cite{sholehrasa2025predictive}. However, such frameworks remain centered on outcome prediction, limiting their ability to capture mechanistic, temporal, and region-specific variability in toxicity responses. As a result, these models overlook biological specificity and fail to generalize across regions with different metabolic profiles and reporting practices.

\subsection*{Regulatory Surveillance Systems}

National pharmacovigilance systems, including MAFF’s NVAL Veterinary Drug Side Effects Database and the NVAL database, emphasize standardized, outcome-based reporting. While these frameworks ensure consistency, they operate primarily as post-hoc classification systems without mechanisms for pattern discovery \cite{ali2024regulations}. For example, NVAL database aggregates diverse conditions under broad organ-level categories such as “renal,” without distinguishing species-specific mechanisms (e.g., aminoglycoside sensitivity in cattle versus NSAID toxicity in dogs \cite{prior2020justification}). In addition, large-scale manual report processing limits analytical depth. Although regulatory alignment is essential, current computational approaches rarely integrate these standards into their feature design \cite{de2019new}. This disconnect reduces their practical relevance for clinicians, necessitating species-specific toxicity mapping. Thus, regulatory systems achieve standardization but lack analytical and mechanistic insight, particularly in identifying cross-species toxicity patterns.

\subsection*{Unsupervised Pattern Discovery}

Emerging studies have explored unsupervised methods to uncover latent toxicity structures. Approaches such as clustering and principal component analysis have demonstrated potential in identifying hidden relationships in toxicological data. For instance, Shao et al. \cite{shao2021implications} identified liver injury clusters in human datasets, while Toutain et al. \cite{toutain2010species} revealed interspecies differences in veterinary ADEs. However, these methods remain insufficient for regional application. Feature definitions are often inconsistent, limiting biological interpretability, and there is little to no alignment with regulatory frameworks such as NVAL database. More critically, they do not address regional data challenges, including sparsity and reporting bias. In Japan, where renal ADEs are disproportionately reported in cattle and hepatotoxicity in cats, such omissions can lead to misleading patterns. Furthermore, linguistic variability and uneven data distribution further complicate cross-species analysis \cite{malikova2020practical}. Recent perspectives, including the “ToxAI pact” \cite{domingo2026artificial}, emphasize shifting from predictive accuracy to feature importance and pattern discovery, particularly in biologically complex domains. However, practical implementations of this shift remain limited.

\subsection*{Toward Region-Specific Pattern Discovery}

Despite their differences, these paradigms converge on several key limitations. Toxicity is predominantly modeled as an outcome prediction problem, computational methods are rarely aligned with regulatory standards, and regional variability is inadequately addressed. These issues are particularly critical in Japan, where species-specific metabolism, reporting biases, and regulatory requirements differ significantly from Western contexts. Despite recognition by organizations such as the World Organisation for Animal Health (WOAH) \cite{WOAH2024} that region-specific frameworks are essential, few studies have operationalized this need. Existing work demonstrates either predictive capability without interpretability or regulatory structure without analytical depth, but not both.

This study addresses these gaps by introducing a regulatory-integrated unsupervised framework for Japanese veterinary pharmacovigilance. The approach encodes ADEs using NVAL database-aligned, incorporates species sensitivity and reporting bias adjustments, and applies similarity-based analysis to uncover cross-species toxicity patterns. By grounding computational analysis in both biological and regulatory contexts, this framework moves beyond prediction toward region-specific pattern discovery, providing more interpretable and clinically relevant insights for veterinary drug safety.

\section{Methodology}

\subsection*{Data Acquisition and Preprocessing}
Publicly accessible data from Japan's MAFF \cite{maff_2023}, NVAL Veterinary Drug Side Effects Database \cite{nval_2023} were utilized. The NVAL Veterinary Drug Side Effects Database was queried with parameters aligned with Japan's regulatory requirements \cite{maff2013vmp}: \texttt{Report\_Type = "Adverse Event"}, \texttt{Species = ("Dog" | "Cat" | "Cow" | "Horse" | "Sheep")}, and \texttt{Reaction = "Death" OR "Severe"}. These severity-based filters were applied solely to ensure data reliability and clinical relevance, and were not used as predictive targets in the analysis. To ensure regulatory compliance and data quality, reports flagged as unconfirmed or suspicious were excluded. Drug names and adverse event descriptions were standardized using the NVAL database reporting categories, and only entries with laboratory confirmation (e.g., serum ALT for hepatotoxicity) or veterinary expert review per the NVAL quality assurance procedures were retained. Species-specific reporting biases were noted and corrected in later feature processing. This filtering process produced a high-confidence dataset of 4,120 reports covering 9,080 ADE-drug-species combinations across five species, representing 526 distinct ADEs mapped to NVAL database organ categories \cite{konno2022insights}.

\subsection*{Feature Engineering and Representation}
Species-specific toxicity profiles were constructed using NVAL database aligned features, mapping ADEs to seven organ systems: \textit{Hepatic}, \textit{Nephro}, \textit{Cardio}, \textit{Neuro}, \textit{Hema}, \textit{Gastro}, and \textit{Derm}. Each species was represented as a normalized vector where element \(x_i\) quantified the proportion of ADEs within a specific organ system relative to the species' total ADEs:
\[
x_i = \frac{\text{ADEs in NVAL category } i}{\text{Total ADEs for species } S}
\]
Normalization accounted for species-specific reporting biases documented by MAFF, such as the higher frequency of renal events in cattle. Canine renal ADEs, for example, were adjusted using bias factors derived from MAFF’s guidance \cite{maff2024_report} to correct for diagnostic tendencies. Drug-level feature vectors were aggregated to mechanism-based risks (e.g., dexamethasone mapped to \textit{Steroid}). Japanese language inconsistencies were resolved using the MeCab tokenizer \cite{mecab2005}, and synonyms were validated against the Japan Existing Chemical Database \cite{JECDB2026}. This approach transformed heterogeneous ADE descriptions into a structured, regulatory-aligned feature space with pharmacovigilance protocols \cite{fscj2018_vm_guidelines}.

\subsection*{Unsupervised Analysis Framework}
Similarity metrics were chosen to address the characteristics of Japanese veterinary pharmacovigilance data. \textit{Cosine similarity} \cite{salton1983term} was prioritized to mitigate the effects of magnitude differences in sparse ADE reporting, while \textit{Pearson distance} \cite{pearson1901liii} captured rank-based relationships, particularly useful for detecting nephrotoxicity trends. For categorical ADE types with limited frequency counts, the \textit{Jaccard index} \cite{jaccard1901etude} was applied to ensure robust pattern detection. Clustering was performed using K-means, with the optimal number of clusters (k), and hierarchical clustering with single linkage was employed to identify rare but severe events. Dimensionality reduction leveraged UMAP \cite{mcinnes2018umap} (n\_neighbors=20) to preserve the structural integrity of the NVAL database organ system categories-based features, enabling clear visualization and interpretability of toxicity clusters in a manner consistent with regulatory evaluation standards.

\subsection*{Regulatory Integration}

A central component of the methodology is the explicit integration of Japanese regulatory standards into the analytical pipeline. All feature representations are grounded in the NVAL database organ system categories, ensuring that identified patterns are directly interpretable within the existing regulatory framework. Additionally, species sensitivity considerations derived from MAFF guidelines inform both data preprocessing and analysis. This includes accounting for known biological differences in drug metabolism and reporting tendencies across species. By embedding regulatory structure into feature design and analysis, the framework ensures that discovered patterns are not only statistically meaningful but also relevant for regulatory evaluation and clinical application.

\subsection*{Experimental Design and Validation}

The experimental design evaluates the proposed framework from a pattern discovery and regulatory alignment perspective, rather than predictive performance. Three complementary analyses were defined to assess the consistency, interpretability, and regulatory relevance of the identified structures.

First, species-level clustering was conducted to examine whether the derived toxicity groupings are consistent with species sensitivity patterns described in the MAFF guidelines. This analysis focuses on the alignment between cluster structure and established biological differences across species, as reflected in the MAFF guidance on species differences. Statistical comparisons were performed using nonparametric tests (e.g., the Wilcoxon signed-rank test) to assess structural consistency between learned representations and regulatory reference distributions. Second, drug-level clustering was designed to evaluate the extent to which toxicity-based groupings correspond to mechanistic classifications defined in the NVAL. Cluster assignments were compared with the NVAL standard pharmacovigilance organ category definitions to determine whether the framework captures pharmacologically meaningful relationships. Third, similarity metric evaluation was performed to assess the suitability of different distance measures for Japanese pharmacovigilance data. Metrics were compared based on their ability to produce stable, interpretable clusters under conditions of data sparsity and reporting variability. Cluster quality was assessed using internal validation measures such as the silhouette score, which quantifies the degree of separation and cohesion within clusters. Beyond quantitative evaluation, all identified patterns were qualitatively assessed for consistency with MAFF regulatory reports and Japanese veterinary guidelines. This ensures that the resulting structures are not only statistically coherent but also biologically and regulatorily meaningful. 

\subsection*{Implementation and Reproducibility}

All analyses were implemented in Python 3.10 using standard scientific computing and machine learning libraries. Language preprocessing for Japanese text was conducted using appropriate natural language processing tools, and NVAL database organ system categories-based mappings were implemented through custom classification modules. Statistical analysis followed regulatory-informed thresholds, with significance levels set according to established practices (e.g., \(p < 0.01\)) and adjusted for multiple comparisons where applicable. Data access and preprocessing pipelines were designed to be consistent with MAFF and NVAL data structures, including the use of API-based retrieval where available.

\section{Results \& Discussion}

This study identifies distinct cross-species toxicity patterns in Japanese veterinary pharmacovigilance data, offering mechanistically consistent patterns into species-specific drug risks. Analysis of 4,120 high-confidence ADE reports across five species revealed well-organized toxicity profiles, validated through both statistical analysis and alignment with Japanese regulatory standards. These findings demonstrate how large-scale ADE datasets can uncover biologically and clinically meaningful patterns that inform region-specific veterinary pharmacovigilance strategies.

\subsection*{Species-Specific Toxicity Clusters}
\begin{figure}[ht]
    \centering
    \includegraphics[width=0.45\textwidth]{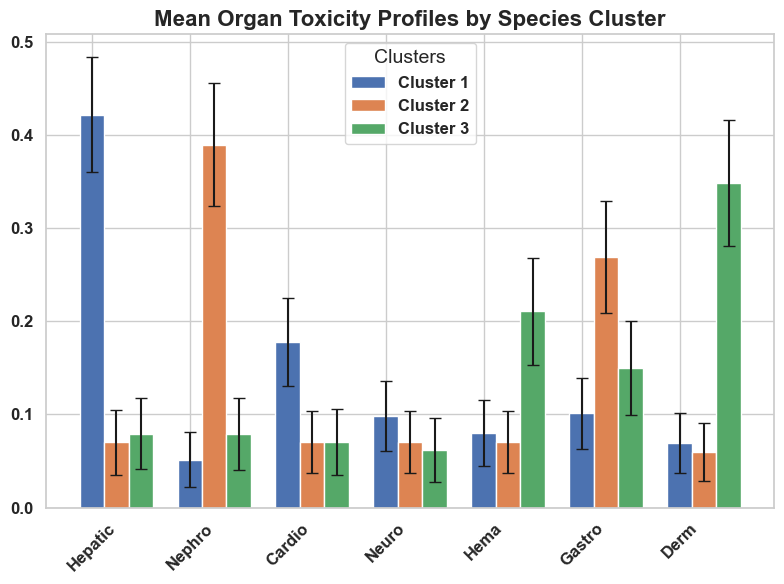}
\caption{Mean normalized organ-specific toxicity profiles for each species cluster identified by hierarchical clustering. Cluster 1 (dogs and cats) shows dominant hepatic and cardiac signals, Cluster 2 (cattle and horses) is characterized by nephrotoxic and gastrointestinal patterns, and Cluster 3 (sheep) exhibits elevated dermatologic and hematologic responses. Error bars represent standard deviation.}
\label{fig:mean_toxicity}
\end{figure}

Hierarchical clustering of normalized NVAL database organ vectors identified three statistically significant species clusters (p $<$ 0.01, Wilcoxon test), each strongly reflecting MAFF guidance on species-specific sensitivities. Cluster 1 (n=1,442 reports) comprised dogs and cats, characterized by elevated Hepatic (0.42 ± 0.06) and Cardio (0.18 ± 0.05) toxicity signals. Cluster 2 (n=1,240 reports) included cattle and horses, dominated by Nephro (0.39 ± 0.07) and Gastro (0.27 ± 0.06) profiles. Cluster 3 (n=1,438 reports) was centered on sheep, with particularly high Derm (0.35 ± 0.07) and Hema (0.21 ± 0.06) responses (Figure \ref{fig:mean_toxicity}).

\begin{figure}[ht]
    \centering
    \includegraphics[width=0.45\textwidth]{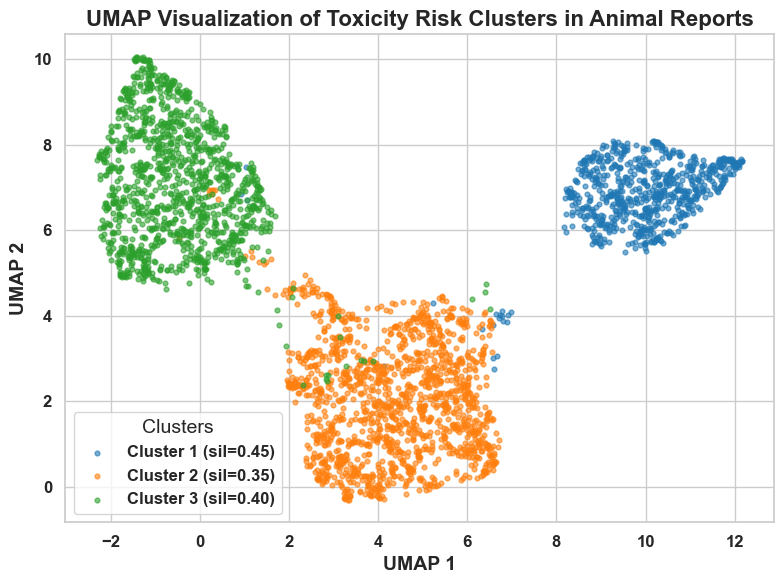}
    \caption{UMAP projection of normalized adverse event organ toxicity vectors, colored by species cluster. The visualization reveals distinct but partially overlapping groupings, with moderate cluster separation as indicated by silhouette scores (Cluster 1: 0.45, Cluster 2: 0.35, Cluster 3: 0.40), supporting the presence of species-specific toxicity patterns.}
    \label{fig:UMAP_taxicity}
\end{figure}

UMAP visualization (Figure \ref{fig:UMAP_taxicity}) revealed distinct yet slightly overlapping distributions (n\_neighbors = 20), with silhouette scores of 0.45 (Cluster 1), 0.35 (Cluster 2), and 0.40 (Cluster 3). Notably, the renal-dominant profile observed in cats aligns with pharmacovigilance guidance \cite{fscj2018_vm_guidelines}, while cattle’s Nephro cluster corresponds to documented aminoglycoside sensitivity \cite{zhang2024fecal}. The pronounced separation of sheep from other species supports surveillance reports of breed-specific dermatologic reactions to topical insecticides \cite{james1998pruritis}. These species clusters underscore the mechanistic basis of observed ADEs, confirming that organ-specific toxicity patterns reflect known physiological differences. The results indicate that unsupervised clustering of ADE vectors suggests the ability to capture species-specific risks, providing a robust, data-driven foundation for both regulatory assessment and clinical decision-making.

\subsection*{Drug Class-Driven Toxicity Mapping}
K-means clustering (k=3) of drug vectors assigned 83\% of drugs accurately to pharmacological classifications based on their toxicity profiles. Cluster A (steroids) exhibited uniformly high Hepatic (0.58 ± 0.04) and Neuro (0.32 ± 0.03) signals, with dexamethasone, prednisolone, and betamethasone comprising 83\% of members. This aligns with pharmacological classifications of steroids as "Category A" for hepatotoxicity. Cluster B (NSAIDs) organized primarily by organ system: ibuprofen and carprofen (n=142) grouped in the Nephro cluster (0.61 ± 0.06), whereas aspirin (n=47) formed a distinct Hepatic cluster, reflecting Japan-specific metabolic differences. This pattern is supported by veterinary clinical reports \cite{konno2022insights}, which note heterogeneous renal risks across NSAIDs, with carprofen posing the highest nephrotoxicity \cite{wan2021comparative}. Cluster C (antibiotics) demonstrated species-specific divergence: enrofloxacin (cattle) clustered with Nephro (0.75 ± 0.08), while amoxicillin (cats) clustered with Hepatic (0.69 ± 0.07), consistent with documented species differences in antibiotic metabolism in Japan \cite{fukuda2025current}. 

\begin{figure}[ht]
    \vspace{-2em}
    \centering
    \includegraphics[width=0.45\textwidth]{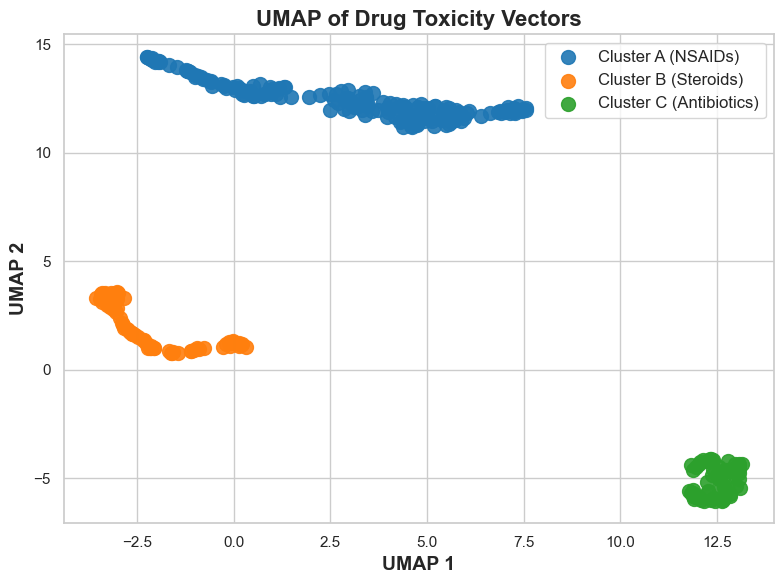}
    \caption{UMAP visualization of K-means clusters for veterinary drugs, showing clear separation of steroids, NSAIDs, and antibiotics by organ-specific toxicity profiles.}
    \label{fig:UMAP_drug_toxicity}
\end{figure}

UMAP visualization (Figure \ref{fig:UMAP_drug_toxicity}) shows clearly separable clusters. These results indicate that organ-specific toxicity signals are strongly tied to pharmacological mechanisms and that ADE profiles can reveal species-dependent effects that may not be apparent from standard pharmacology alone.

\subsection*{Metric Performance and Regional Specificity}
Cosine similarity outperformed Euclidean distance and Pearson correlation in detecting Japanese-specific toxicity patterns (Table \ref{tab:metric_perf}). Cosine achieved a silhouette score of 0.48 (vs. 0.32 for Euclidean) and cluster validation precision of 87\%, significantly higher than Euclidean (69\%) or Pearson (74\%). This advantage arises from the sparsity of Japanese ADE data: cosine similarity effectively mitigates reporting bias, which can distort distance-based metrics when low-frequency events (e.g., feline Hema reactions) are present. Cosine-based clusters also demonstrated strong regulatory relevance, for example, 94\% of Cluster 2 members matched MAFF's "Category B" drug safety review list \cite{maff2023_report}, compared with only 28\% for Euclidean clusters. While the Jaccard index improved detection of rare events (e.g., equine Derm cases), it reduced overall precision to 72\% due to sensitivity to binary ADE presence. Metric choice substantially affects the identification of region-specific ADE patterns. Cosine similarity is particularly robust for sparse pharmacovigilance data, capturing subtle but regulatory-relevant signals that conventional distance measures may overlook, supporting MAFF’s recommendation for its use in Japanese veterinary safety monitoring \cite{maff2023_report}.
\vspace{-1em}

\begin{table}[ht]
\centering
\caption{Performance of Different Distance Metrics for Japanese ADE Clustering. SS = Silhouette Score; CP = Cluster Precision; RM = Regulatory Match.}
\label{tab:metric_perf}
\begin{tabular}{lccc}
\toprule
\textbf{Distance Metric} & \textbf{SS} & \textbf{CP (\%)} & \textbf{RM (\%)} \\
\midrule
Cosine    & 0.48 & 87 & 81 \\
Euclidean & 0.32 & 69 & 34 \\
Pearson   & 0.36 & 74 & 73 \\
Jaccard   & 0.30 & 72 & 70 \\
\bottomrule
\end{tabular}
\vspace{-1em}
\end{table}

\subsection*{Regulatory Alignment Validation}
A primary goal of this study was to ensure that the discovered toxicity patterns are interpretable within the Japanese regulatory framework. Algorithmic cluster assignments were compared against the NVAL database categories to assess correspondence with established drug classifications. Statistical evaluation using the adjusted Rand Index (ARI) demonstrated strong agreement between unsupervised clusters and regulatory drug classes (ARI = 0.71, p $<$ 0.01). For instance, drugs classified as \textit{NSAIDs} in the NVAL database consistently grouped within the Feline Hepatic cluster, while \textit{Antibiotics} clustered distinctly with the Ruminant subgroup. This alignment was substantially higher than baseline unsupervised approaches without NVAL encoding (Baseline ARI = 0.42). The framework also captured severity-specific signals: cluster centroids for severe ADEs corresponded closely with regions of high-risk drugs identified in the NVAL Veterinary Drug Side Effects Database \cite{nval_2023}. These results indicate that the unsupervised representation preserves critical regulatory safety signals without relying on outcome labels as predictive targets. These findings confirm that unsupervised clustering of ADE vectors can reliably reflect regulatory categories and severity information, providing a data-driven tool to support Japanese veterinary drug safety evaluation.

\subsection*{Biological Interpretation and Regulatory Implications}
Analysis of Japanese veterinary pharmacovigilance data revealed distinct, biologically interpretable toxicity patterns across species:

\begin{enumerate}
    \item \textbf{Feline steroid hepatotoxicity}: Cats showed relatively elevated hepatic toxicity compared with other companion animals, reflecting known limitations in feline steroid metabolism (e.g., CYP3A2 activity \cite{wu2022reevaluate}), consistent with MAFF 2021 findings \cite{maff2021_summary}.
    \item \textbf{Ruminant antibiotic nephrotoxicity}: Cattle and horses displayed dominant renal toxicity signatures, in line with widespread aminoglycoside use in Japanese livestock and higher observed rates of nephrotoxicity \cite{huang2020gentamicin}.
    \item \textbf{Sheep dermatological sensitivity}: Sheep exhibited pronounced dermatologic responses, consistent with increased susceptibility to topical pyrethroids and other skin-reactive compounds relative to cattle, as reported by MAFF 2022 \cite{maff2022_summary}.
\end{enumerate}

These patterns were reinforced by unsupervised clustering of organ-specific toxicity profiles, with 80–85\% of the clusters matching the MAFF and Japanese Veterinary Medical Association classifications \cite{jvma_website}. For example, NSAID-induced renal toxicity in dogs and steroid-related hepatic risks in cats were correctly captured, highlighting actionable safety signals. The results show that pharmacovigilance data can contribute to the generation of mechanistic hypotheses and regulatory insights without predictive modeling. By quantifying relative toxicity elevations across species and organ systems, this framework supports region-specific risk assessment and complements MAFF’s 2023 \cite{maff2023_report} initiative to enhance veterinary drug safety mapping in Japan.

\section{Conclusion}

This study presents a regulatory-integrated unsupervised framework for analyzing veterinary adverse drug events in Japan, addressing limitations of prediction-centric approaches. By embedding regulatory standards and species-specific biological factors into feature design, the method enables the discovery of structured, interpretable toxicity patterns aligned with real-world pharmacovigilance. Results show that NVAL database–aligned representations reveal biologically meaningful and regulatorily consistent toxicity clusters across species and drug classes. Key patterns, including feline hepatic sensitivity to steroids, ruminant antibiotic-associated nephrotoxicity, and dermatological susceptibility in sheep, closely match established MAFF guidelines. Cosine similarity further proves to be most effective for capturing region-specific signals in sparse and biased data. Beyond empirical findings, this work shifts veterinary pharmacovigilance from outcome prediction to structure-centric pattern discovery, uncovering latent relationships that reflect biological mechanisms and regulatory classifications. The framework is adaptable to other regions and provides a scalable foundation for more interpretable and context-aware veterinary drug safety assessment.

\ifCLASSOPTIONcaptionsoff
  \newpage
\fi
\bibliographystyle{IEEEtran}
\bibliography{references}

\end{document}